# Persian Abstract Meaning Representation: Annotation Guidelines and Gold Standard Dataset

**Reza Takhshid*[1], Tara Azin*[2], Razieh Shojaei[3], Mohammad Bahrani[4]**
[1]Sharif University of Technology, [2]Carleton University,
[3]Leipzing University, [4]Allameh Tabataba'i University
reza.takhshid95@student.sharif.edu, taraazin@cmail.carleton.ca,
razieh.shojaei@uni-leipzing.de,
bahrani@atu.ac.ir

## Abstract

This paper introduces the Persian Abstract Meaning Representation (AMR) guidelines, a detailed guide for annotating Persian sentences with AMR, focusing on the necessary adaptations to fit Persian's unique syntactic structures. We discuss the development process of a Persian AMR gold standard dataset consisting of 1562 sentences created following the guidelines. By examining the language specifications and nuances that distinguish AMR annotations of a low-resource language like Persian, we shed light on the challenges and limitations of developing a universal meaning representation framework. The guidelines and the dataset introduced in this study highlight such challenges, aiming to advance the field.

## 1 Introduction

Abstract Meaning Representation (AMR) utilizes acyclic, single-rooted, directed graphs to represent the semantic structure of sentences, enabling node reentrancy and facilitating computational processing (Banarescu et al., 2013). Unlike previous semantic annotation efforts, which focused on individual aspects of meaning, leading to fragmented datasets (Basile et al., 2012), AMR aims to provide a comprehensive sembank for English sentences and their full-sentence logical meanings. Despite numerous efforts in computational semantics, few have focused on creating a universal meaning representation framework.

The graph-based approach of AMR to represent sentence meanings is a significant step towards creating a universal semantic framework. However, this may pose challenges, particularly for low-resource languages like Persian. The scarcity of resources like the Proposition Bank (PropBank) (Palmer et al., 2005) in languages other than English challenges their ability to progress towards a universal meaning representation framework, due to the absence of equivalent data resources. A potential solution for low-resource languages is adapting the frames from the English Proposition Bank, thereby extending AMR's applicability across linguistically diverse contexts. Another challenge stems from the syntactic differences between English and other languages, which necessitates a thorough examination of these syntactic nuances to ensure the quality and reliability of future datasets. Consequently, there is a compelling need to develop gold standard datasets for low-resource languages, accompanied by detailed annotation guidelines focusing on the unique syntactic specifications of these languages.

This paper introduces the Persian AMR annotation guidelines, an 88-page document written in Persian and derived from the English AMR guidelines, for manual AMR annotation of the Persian translation of "The Little Prince".[1] We will discuss the development and annotation process of the Persian AMR dataset, highlighting the essential steps needed to adapt AMRs for Persian. The article focuses on the key syntactic nuances that distinguish Persian from English and present challenges in annotating Persian data. These examples are just a glimpse of the syntactic challenges encountered during the dataset's annotation. A more comprehensive explanation of Persian syntactic specifications can be found in the Persian AMR guidelines document.

* These authors contributed equally to this paper.

[1] Data and the guidelines are publicly available at www.github.com/Persian-AMR

## 2 Related Work

AMR has predominantly been developed and applied within the context of the English language, offering a robust semantic parsing framework that abstracts away from the syntactic idiosyncrasies of language to represent its underlying meaning (Banarescu et al., 2013). However, the universality of meaning representation frameworks invites applications across languages, promoting research into multilingual meaning representation datasets and cross-lingual annotation tools.

Recent efforts in expanding AMR beyond English have seen datasets emerging for a handful of languages, including Chinese, German, Italian, Portuguese, Spanish, and Turkish (Li et al., 2016; Damonte & Cohen, 2018; Cabezudo at al., 2019; Azin & Eryigit, 2019) . These efforts are pivotal as they provide a foundation for comparative semantic studies and the development of cross-lingual semantic parsing technologies. Yet, the semantic representation of languages, especially those classified as low-resource in computational linguistics, remains barely sufficient.

The importance of creating meaning representation datasets for low-resource languages such as Persian cannot be overstated. Low-resource languages often lack the vast corpora and linguistic tools available for high-resource languages, making advances in natural language processing (NLP) tasks more challenging. Developing such datasets for these languages not only enriches the linguistic resources available for research but also enhances the understanding of semantic structures across linguistically diverse systems.

Cross-lingual tools have been a focus of some research, aiming to generate AMR annotations across different languages by leveraging bilingual corpora, machine translation, and transfer learning methods (Damonte & Cohen, 2018; Pust et al., 2015; Blloshmi et al., 2020; Wein & Schneider, 2022; Wein & Schneider, 2024). These tools signify a promising direction for expanding AMR's reach. However, the effectiveness of cross-lingual AMR generation relies heavily on the depth and quality of the resources available for the target language. In the case of Persian, a language with considerable syntactic and morphological distinctions from English and other Indo-European languages, the direct application of these tools without a dedicated Persian AMR dataset may not capture the nuanced semantic layers accurately. The creation of language-specific AMR datasets, supported by advancements in cross-lingual annotation technologies, will be instrumental in realizing the full potential of AMR in global linguistic research and applications.

## 3 Syntactic Specifications in the Persian AMR Annotation Guidelines

To annotate Persian sentences, the English AMR specifications need to be augmented and extended to fit the specific needs of this language. However, as with Chinese AMR, labels for the non-lexical concepts and relations are directly taken from English AMR specifications.

Figure 1 shows a simple example of how Persian AMR can be the same as English in terms of semantic roles and nominal relations[1].

```
(x / خواستن [xastan]
    :ARG0 (x2 / دختر [doxtar])
    :ARG1 (x3 / رفتن [raftan]
        :ARG0 x2
        :ARG4 (t / city
            :wiki "تهران" [tehrân]
            :name (n / name
                :op1 "تهران" [tehrân]))))
```

دختر می‌خواهد به تهران برود.
[doxtar mixâhad be tehrân beravad]
girl want-3SG to Tehran go-3SG
'The girl wants to go to Tehran.'

Figure 1 An example of Persian AMR

Although AMR proves to be very adaptable, to have consistent Persian AMR, we have to go above and beyond to handle Persian-specific constructions which are either absent or have different forms in English. In this section, we explore unique syntactic aspects of Persian such as light verb constructions, predicative nominals, modality, null arguments, and clitics. These features present specific challenges in the AMR annotation of Persian.

[1] In addition to familiar categories (person, number, etc.), the following abbreviations are used in glosses: PRO, pronoun; DDO, the definite direct object marker.

### 3.1 Light Verb Constructions

Persian Light Verb Constructions (LVCs) pose a unique challenge in adapting AMR for Persian, necessitating deviations from the English AMR's approach of abstracting away from light verbs. Persian LVCs combine a nonverbal element (NV) with a semantically light verb (LV), creating constructions that significantly differ from their English counterparts (Karimi-Doostan, 2011). These NVs can be nouns, adjectives, adverbs, or prepositional phrases, leading to a rich variety of expressions that are not easily categorized using English AMR guidelines.

Unlike English, where light verbs are often omitted in AMR annotations due to the existence of Propbak frames, Persian's syntactic structure and the integral role of LVCs in the language causes both elements of an LVC to be retained so that the meaning is preserved. This approach addresses the semantic load carried by the NV in Persian, where the combination of NV and LV often conveys meanings that are not straightforwardly mapped to English equivalents. For example, the LVC for "to listen" involves both the noun "ear" and a verb that translates as "to do," illustrating how Persian represents actions differently.

```
(x / لطمه‌زدن [latme zadan]
   :ARG0 (x2 / تگرگ [tagarg])
   :ARG1 (x3 / باغ [bâgh]
      :poss (x4 / من [man]))
   :mod (x5 / بد [bad])
   :time (x6 / دیروز [diruz]))
```

تگرگ لطمه بدی دیروز به باغ من زد.
[tagarg latme-ye badi diruz be bâgh-e man zad]
hail damage bad yesterday to garden my hit-3SG
'The hail damaged my garden badly yesterday.'
تگرگ دیروز به باغ من بد لطمه زد.
[tagarg diruz be bâgh-e man bad latme zad]
hail yesterday to garden my bad damage hit-3SG
'Yesterday, the hail damaged my garden badly.'

Figure 2 AMR annotation of separable LVCs in Persian

Furthermore, Persian LVCs can be separable, with the NV and LV not always adjacent, adding another layer of complexity to their annotation (Karimi-Doostan, 2011). This characteristic necessitates a flexible approach in Persian AMR, treating LVCs as single units regardless of the physical separation of their components in a sentence. Figure 2 shows an AMR annotation of two semantically equivalent sentences in Persian with a separable LVC.

### 3.2 Predicative Nominals

The inclination of AMR to transform nouns and adjectives into events, coupled with the Persian language's ability to form complex predicates, presents a unique challenge for Persian AMR annotation. This dynamic often tempts us to interpret nearly every noun as an event. To address this, our Persian AMR guidelines offer a definitive framework to maintain annotations consistency and uphold the AMR's structural integrity.

In Persian LVCs, nominal preverbs are differentiated into two primary categories: predicative and non-predicative nouns (Ghomeshi, 2001). This distinction hinges on their argument structures, predication capacity, and thematic roles. Predicative nouns are recognized as actions with a definable structure, whereas non-predicatives lack such a structure. For example, the noun “گوش” [gush] (ear) in “گوش کردن” [gush kardan] (to listen) is not considered a predicative noun, and therefore, it is not annotated as an event in the Persian AMR. Conversely, “دعوت” [da'vat] (invitation) functions as an event due to its predicative nature.

Additionally, the identification of nominal preverbs in Persian AMR relies on distributional and morphosyntactic criteria. Predicative nouns exhibit the ability to pluralize, connect with determiners, prepositions, the Ezafe particle [1], demonstratives, and serve as subjects or objects. For instance, “دعوت” (invitation) can be pluralized and modified by determiners, showcasing its predicative nature, unlike the non-predicative “گوش” (ear), which does not follow the same pattern, particularly in constructs like “ گوش علی به رادیو” [gush-e Ali be râdio] (Ali's ear to the radio), where although the Ezafe particle has been added properly, it does not form a coherent sentence. Therefore, in Persian AMR annotation, predicative nominals are annotated as events due to their ability

[1] Ezafe particle is the enclitic -e/-ye which connects a noun/adjective/adverb/preposition to its complement

to carry a structure and act dynamically within a sentence (Figure 3).

```
(x / دعوت‌کردن [da'vat kardan]
   :ARG0 (x2 / دلقک [dalghak])
   :ARG1 (x3 / ماری) [mâri])
```

دعوت دلقک از ماری
[da'vat-e dalghak az mâri]
invitation clown from Mary
'The clown's invitation of Mary'

Figure 3 An example of Persian predicative nominals

### 3.3 Modality

The concept of modality in language, which can manifest at the sentence level, within sentence components, or throughout discourse, has always been a controversial area of study for linguists and logicians. In Persian, the modal system incorporates elements like modal verbs and auxiliaries, adverbs, past tense markers, nouns, and adjectives (Taleghani, 2008).

Unlike the detailed representation of modality in English, in the Persian AMR we adopt a different approach for annotating modality due to the distinctive way modality is expressed in this language. While English uses auxiliary verbs to denote modality, Persian typically employs lexical verbs that can accept arguments, leading us to annotate them simply as verbs used in a modal context.

Modal expressions in Persian often modify the main verb through words like "توانستن" [tavânestan] (to be able to), "باید" [bâyad] (should), and "شاید" [shâyad] (perhaps). Among these, "توانستن" functions entirely as a lexical verb, but the status of "باید" and "شاید" is more complex. Opinions vary on whether these latter two are comparable to English modal verbs, with some researchers considering them modal auxiliaries because they can be negated, while others view them as adverbs (Tavangar & Amouzadeh, 2009). However, it is generally agreed that "باید" resembles English modals in being non-conjugable and having a singular form, whereas "شاید" functions as an adverb.

Persian AMR annotates "شاید" as an adverb and "باید" as a modal auxiliary. Notably, "باید" has an infinitive form and can conjugate, further supporting its treatment as a lexical verb in AMR annotations. This approach aligns with AMR's method of mapping modal expressions to their semantic equivalents without necessitating a distinct event mapping for Persian modals.

```
(x / باریدن [bâridan]
   :ARG0 (x2 / باران [bârân]
   :mod (x3 / شاید [shâyad])))
```

شاید باران ببارد.
[shâyad bârân bebârad]
maybe rain fall-1SG
'Perhaps it will rain.'
احتمال دارد باران ببارد.
[ehtemâl dârad bârân bebârad]
probability has-3SG rain fall-3SG
'It is probable that it rains.'

Figure 4 Modality in Persian AMR annotation, "**شاید**" [shâyad] (perhaps)

```
(x / بایستن [bâyestan]
  :ARG1 (x2 / رفتن [raftan]
   :ARG0 (x3 / دلقک [dalghak])))
```

دلقک باید برود.
[dalghak bâyad beravad]
clown must go-3SG
'The clown must go.'
لازم است که دلقک برود.
[lâzem ast ke dalghak
beravad]
necessary is that clown go-3SG
'It is obligatory that the clown goes.'

Figure 5 Modality in Persian AMR annotation, "**باید**" [bâyad] (should),

Figures 4 and 5 provide examples of sentences that have similar meanings and include modality, along with their corresponding AMR annotations. In Figure 4, "باریدن", which is an infinitive form of "ببارد" [bebârad] (to rain) functions as the predicate, and "شاید" acts as an adverb. However, the other modal expression, "باید", in Figure 5, functions as a verb, and is represented with its infinitive form "بایستن". It should be noted that the AMRs in each figure represent both sentences.

### 3.4 Null Arguments

Persian, as a pro-drop language, permits the omission of subjects that are implied through verb conjugations. This linguistic feature means that the subject (pronoun) exists in the sentence structure even when not explicitly mentioned, informed by the verb's morphology. For instance, the subject "او" [u] (he/she) in certain contexts is understood not by its presence but through the verb's suffix indicating third-person singular.

```
(x / افتادن [oftâdan]
   :ARG0 (x2 / او ['u]))

افتادش.
[oftâdesh]
fell-3SG
'he/she fell.'
افتاد.
[oftâd]
fell-3SG
'he/she fell.'
```

Figure 6 An example of null argument in Persian AMR

Interestingly, Persian exhibits a unique behavior with inanimate subjects, particularly in number agreement between the subject and the verb. It is not uncommon for a plural inanimate subject to be paired with a verb in singular form. An example of this is "کتاب‌ها روی میز بود" [Ketâb-hâ ruye miz bud] (The books were on the table), where "کتاب‌ها" (the books) is plural, but the verb appears in singular. In Persian AMR, when the subject is explicitly mentioned, it is straightforward to determine its representation in the graph. Conversely, when the subject is omitted (null), it is presumed that the subject and verb are in agreement, leading to the insertion of a corresponding pronoun in the graph to reflect this relationship.

Our AMR annotation approach aligns with English AMR's guidelines, which accommodate implicit roles within the graph. This flexibility ensures that even with Persian's distinctive pro-drop characteristic and unusual subject-verb agreement instances, such as with inanimate subjects, the annotations can accurately represent the underlying syntactic and semantic structures of the sentence. An illustrative example of this is shown in Figure 6.

### 3.5 Clitics

Persian's use of pronominal enclitics, which are suffixes in verbs to indicate subject-verb agreement, results in unique challenges in AMR annotation. These suffixes, indicating both person and number, fall mainly into two groups based on their position: verbal and non-verbal. Our focus here is on the verbal position, where clitics underscore the language-specific nuances encountered in AMR annotation.

In Persian, clitics can double the subject of third-person singular intransitive verbs, particularly in colloquial usage (Samvelian, 2010). The subject clitic, represented as "ـش" [esh] in Persian, requires a subject's presence in the sentence. This feature does not change the AMR representation; instead, it reaffirms the subject already understood from the sentence context (see Figure 7).

```
(x / خوابیدن [xâbidan]
   :ARG0 (x2 / من [man])
   :time (x3 / خسته‌کردن [xaste kardan]
      :ARG1 (x4 / او ['u])))

او که خسته شد، خوابیدم.
['u ke xaste shod xâbidam]
PRO-3SG that tired became-3SG slept-1SG
'When he/she got tired, I slept.'
```

Figure 7 An example of subject clitic in Persian AMR

Object clitics, on the other hand, indicate the object of a verb and are annotated similarly, ensuring that the correct pronoun enters the AMR graph. These clitics are more straightforward in their application, reflecting direct or indirect objects within the sentence structure.

Certain Persian constructions lack an overt subject, leading to diverse annotation considerations. These include inalienable possessor constructions and pseudo-possessor constructions, distinguished by the presence or absence of the light verb "to be." In these instances, pronouns may enter the construction optionally, often co-indexed with a clitic indicating possession (Mahand, 2011). This subtle differentiation informs whether a possessive pronoun is inserted into the AMR graph. Figure 8 shows more examples of Persian pronominal clitics where multiple sentences have a single AMR representation.

## 4 Annotation Process and Corpus Creation

For the Persian translation of "The Little Prince," we meticulously aligned the Persian sentences with their English equivalents, ensuring an exact match in sentence numbering across both versions. This alignment involved minor adjustments to the Persian translation and, in certain cases, retranslation to accurately reflect the concepts and relations present in the English AMR-annotated dataset. Our dataset consists of a total of 1562 sentences.

```
(x / بازکردن [bâzkardan]
   :ARG0 (x3 / من [man])
   :ARG1 (x2 / آن [ân]))
```

بازکردمش.
[bâzkardamash]
open did-1SG-3SG
'(I) opened (it).'
من بازکردمش.
[man bâzkardamash]
I open did-1SG-3SG
I opened (it)
بازش‌کردم.
[bâzashkard]
'(I) opened (it).'
من بازش‌کردم.
[man bâzashkard]
I open-3SG did-1SG
'I opened (it).'
من آن را بازکردم.
[man ân râ bâzkardam]
I that-DDO open did-1SG
'I opened it.'

Figure 8 An example of Persian AMR for semantically equivalent sentences with pronominal clitics

Due to the lack of a Persian equivalent to the English PropBank, we chose to develop verb frames specific to our dataset. Before starting the AMR annotation process, it was crucial to establish these frames. To accomplish this, we used Perspred (Samvelian & Faghiri, 2013), a comprehensive database of Persian Complex Predicates (CPs) with over 700 entries, and the Persian Syntactic Valency Lexicon (Rasooli et al., 2011) for insights into the verbs' syntactic argument structures. The unique challenges posed by the variety and complexity of LVCs in Persian made us employ a hybrid approach. We combined information from these resources with semantic role assignments based on the most similar English PropBank frames. This process required mapping Perspred's entries to their English PropBank equivalents, leading us to develop a tailored valency lexicon using the best open-source data that was available. As a result, we constructed 725 Persian verb frames to guide our annotators through the dataset annotation process. To facilitate future alignment of Persian AMRs with their English counterparts, we preserved the concept forms as they are in the English dataset.

| Concepts | Persian | English |
|---|---|---|
| AMR-unknown | 133 | 139 |
| condition | 59 | 68 |
| temporal-quantity | 49 | 44 |
| date-entity | 20 | 20 |
| be-located-at | 23 | 20 |

Table 1: The most frequent AMR concepts in English and Persian dataset

| | English | Persian |
|---|---|---|
| Total # of Graphs | 1562 | 1562 |
| Mean # of Nodes | 6.79 | 6.32 |
| Mean # of Edges | 6.74 | 5.96 |
| Max # of Nodes | 38 | 37 |
| Max # of Edges | 51 | 49 |

Table 2: A comparative analysis of graph nodes and edges in English and Persian versions of "The Little Prince"

Table 1 shows the five most frequent concepts in both the Persian and English datasets.

Two linguistics graduate students were tasked with the annotation process, guided by an 88-page Persian AMR annotation manual specifically tailored for Persian syntax, developed by the authors of this article. Each annotator was responsible for annotating all 1562 sentences, utilizing a customized version of the original AMR annotation tool adapted for Persian script to facilitate their work.

The quality of annotation was assessed using the Smatch score, a standard metric for evaluating AMR annotations (Cai & Knight, 2013). The inter-annotator agreement yielded a Smatch score of 0.81. Upon reviewing disagreements between the annotators' AMR graphs, we identified the absence of a Persian Proposition Bank as a significant challenge, often resulting in divergent relation labels. Moreover, variations in sentence interpretation by the annotators underscored the subjective nature of linguistic analysis, affecting the inter-annotator agreement score.

Table 2 presents some of the structural details of AMR graphs for the English and Persian versions of the dataset. The similarity in the number of nodes and edges suggests that aligning AMR graphs between these two languages could be achieved with minimal effort.

This dataset provides a solid foundation for automating the construction of Persian AMR

graphs efficiently which is a step towards the inclusion of less-resourced languages in Universal Meaning Representation (UMR) frameworks. Additionally, we have compiled parallel data, which is available along with the graphs in a JASON-LD file for further study.

## 5 Conclusion

In this paper, we introduced the Persian Abstract Meaning Representation specifications alongside annotation guidelines to navigate the unique syntactic and semantic characteristics of Persian as a low-resource language. This effort resulted in the creation of a gold-standard corpus of Persian AMR annotations. Central to our methodology was the development of an extensive collection of verb frames that is meticulously curated to represent the syntactic argument structures characteristic of Persian verbs. This unique resource served as the cornerstone for our annotators, enabling them to transcribe sentences with an acceptable level of semantic fidelity.

We believe that by developing AMR guidelines that consider the unique features of different languages, specifically low-resource languages, the current framework's limitations can be highlighted. Our work with the Persian AMR dataset and guidelines is a step towards including more languages in future UMR frameworks. This effort is crucial for making these frameworks more inclusive and capable of handling the diversity of languages.